# Evaluation of Computational Grammar Formalisms for Indian Languages

Nisheeth Joshi [1], Iti Mathur [2]

[1] [2] Department of Computer Science, Apaji Institute, Banasthali University, Rajasthan, India
*nisheeth.joshi@rediffmail.com [1], mathur_iti@rediffmail.com [2]*

**ABSTRACT**

Natural Language Parsing has been the most prominent research area since the genesis of Natural Language Processing. Probabilistic Parsers are being developed to make the process of parser development much easier, accurate and fast. In Indian context, identification of which Computational Grammar Formalism is to be used is still a question which needs to be answered. In this paper we focus on this problem and try to analyze different formalisms for Indian languages.

## 1. INTRODUCTION

Natural Language Parsing has been an important activity in Natural Language Processing (NLP) development. But, even since the introduction of machine learning techniques into NLP application development, the scenario changed drastically. This new approach appeared to be very promising, as it helped in rapid prototype development of NLP systems. In this technique, large amount of data was used, onto which various models like Hidden Markov Model (HMM), Conditional Random Fields (CRF), Neural Networks (NN), Support Vector Machines (SVM) etc. were applied. These approaches are also termed as statistical approaches or Statistical Natural Language Processing (SNLP). This was a very effective way of application development, with applications attaining 60-75% accuracy with very little effort. Unfortunately, this approach soon lost its shine as after a point of optimized performance, they become very less helpful in improvement of the systems[1]. Moreover, it failed to implement broad coverage parsing or deep parsing.

Due to this reason, NLP researchers, in order to improve performance of their systems, tried a new approach. They initially stared with a rule based (traditional) approach. This was called the seed data. Once this was done, it was then supplied to machine learning techniques. This approach was termed as hybrid approach (partially rule based and partially statistical). We can find evidence of improved systems in literature which used this approach[2][3]. This approach even helped in development of probabilistic parsers like Stanford parser[4], Charniak parser[5], MaltParser[6]. These all parsers where supplied with different computational grammars formalisms or with treebanks, which were developed using manually parsed sentences, based on one of the formalisms, for example Penn Treebank[7] TIGER Treebank[8] Paraguay Dependency Treebank[9]. In one or the other way, grammar formalisms were used for development of deep parsers. In this paper, we attempt to study the performance of some of the popular computational grammar formalism techniques, which could be used in development of deep language processing applications like a deep parser, machine translators, semantic role labeler etc.

The motivation for this study came from the fact that free word order is one of the areas were grammar formalism has not yet reached the level of good accuracy. In this area there have been numerous claims to prove superiority of dependency grammar over other formalisms[10][11]. But, often discussions based on this formalism ignore more practical aspects like usability and expressivity. In order to examine free word order approach, we conducted our study on Hindi. Since all other Indian languages follow the same phenomena. The approach suggested in this study can be applied to other languages.

## 2. COMPUTATIONAL GRAMMAR FORMALISMS

In general computational grammars can be divided into three categories based on their functionality. They are Phrase Structure Grammars, Dependency Grammars, and Hybrid Grammars. A phrase structure grammar is the one which uses the approach shown by transformational grammars where specific tree positions are associated with assignments of various syntactic roles, such as subject and object. This concept is the motivation for having elements appear in various positions in the tree in the process of deriving the final syntactic structure. Some of the popular formalisms of this category are Tree Adjoining Grammar (TAG) [12], Head Driven Phrase Structure (HPSG) [13] Grammar. Hybrid grammar augments phrase structure grammar by expressing non-projective syntactic relations, while maintaining a more formally defined architecture then phrase structure grammar. Lexical Functional Grammar (LFG)[14] is an example of hybrid grammar.

A dependency grammar consists of a set of words and a set of directed binary dependency relations between words, such that
- No words depends on itself
- Each dependent has one and only one head
- A head may have many dependents
- There is one distinguished word which is the head of the sentence and depends on no other word
- All other words in a sentence are dependents such that the whole sentence is connected.

Dependency grammars have been studied by Gaifman[15] who studied linear precedence in dependency relations, Hudson[16] who introduced word grammar, Starosta [17] who studied lexicase grammar and Bharti et al[18] who showed the similarities between Paninian Grammar (PG) and dependency grammar and the suitability of PG in Indian context.

### 3. METHODOLODY

In order to understand the pros and corns of different grammar formalism, we tested all three types of grammar formalisms. From phrase structure stable, we selected TAG, LFG from hybrid and PG from dependency framework.

We developed parallel grammars for all three frameworks and took a detailed note of development process and variations in syntactic structures. We recorded time taken to construct each sentence, the total time taken to complete the task and the average time taken for the task. We also noted the difficulty level with which each grammar was developed.

Since all three grammar formalism are somewhat distinct in nature, it was very much necessary to develop a mechanism which would not be biased towards one grammar and penalize others. To ensure the equivalence, we tested each grammar using the same set of sentences. The test case contained grammatical as well as ungrammatical sentences. Each grammar was required to distinguish between the two categories. Moreover each grammar was required to provide predicate, arguments and modifiers for each sentence which was parsed. Figure 1 and 2 give a brief idea of the type of sentences used. Various types of sentences used in the test case were:
- Basic sentences with auxiliary verbs
- Sentences having case assigning post positions
- Sentences marking subjects
- Adpositional sentences
- Sentences with generative constructions
- Sentences with descriptive adjectives
- Sentences with predicative adjectives
- Sentences with relative/co-relative constructions

```
(a) लड़की ने लड़के को मारा
    ladkii ne ladke ko mara
    girl-Erg boy-Acc hit
(b) मारा लड़की ने लड़के को
    mara ladkii ne ladke ko
(c) लड़के को लड़की ने मारा
    ladke ko ladkii ne mara
(d) मारा लड़के को लड़की ने
    mara ladke ko ladkii ne
(e) लड़के को मारा लड़की ने
    ladke ko mara ladkii ne
(f) लड़की ने मारा लड़के को
    ladkii ne mara ladke ko
```
Figure 1: Simple Hindi Test Sentence with various variations [19]

```
(a) जो खड़ी है वो लड़की लंबी है
    jo khari hai vo ladkii lambii hai
    Rel standing be Co-Rel girl tall is
(b) जो लड़की खड़ी है वो लंबी है
    jo ladkii khari hai vo lambii hai
(c) वो लड़की लंबी है जो खड़ी है
    vo ladkii khari hai jo lambii hai
(d) * वो लड़की लंबी है जो लड़की खड़ी है
    vo ladkii lambi hai jo ladkii khari hai
(e) * वो लंबी है जो लड़की खड़ी है
    vo lambi hai jo ladkii khari hai
(f) वो लड़की जो खड़ी है लंबी है
    vo ladkii jo khari hai lambii hai
(g) * वो जो लड़की खड़ी है लंबी है
    vo jo ladkii khari hai lambii hai
```
Figure 2: Simple Hindi Test Sentence with various variations [19]

LFG and DG had no problems to handle these type of sentences. We used Lexicalized TAG (LTAG) which is the modification of TAG and can handle word order variation.

We did this study using ten grammar writes, which were provided with the sentences and were asked to construct the grammars for each sentence. In order to understand the usability of each grammar, we provided each writer with 35 sentences from various categories, as discussed above. Each writer was provided with a short tutorial of each grammar. Shortly after the tutorial of a particular grammar, the writers were asked to implement the sentences for the said grammar. Their performance was calculated on the measure discussed above.

| Writer | PG | | TAG | | LFG | |
|---|---|---|---|---|---|---|
| | Dur. | Acc. | Dur. | Acc. | Dur. | Acc. |
| W1 | 9 min | 74% | 28 min | 71% | 41 min | 61% |
| W2 | 8 min | 80% | 31 min | 84% | 43 min | 73% |
| W3 | 8 min | 79% | 27 min | 83% | 48 min | 64% |
| W4 | 10 min | 60% | 35 min | 88% | 56 min | 66% |
| W5 | 13 min | 98% | 36 min | 88% | 61 min | 73% |
| W5 | 17 min | 93% | 37 min | 94% | 67 min | 71% |
| W7 | 9 min | 83% | 38 min | 97% | 71 min | 79 % |
| W8 | 11 min | 87% | 39 min | 93% | 59 min | 67% |
| W9 | 11 min | 95% | 35 min | 89% | 55 min | 69% |
| W10 | 7 min | 70% | 34 min | 73% | 59 min | 62% |
| Average | 10.3 min | 81.9 % | 34 min | 86.0 % | 56 min | 68.5 % |

Table 1: Total Time Taken and Accuracy Achieved by Each Grammar Writer

| Task | Difficulty | | |
|---|---|---|---|
| | PG | TAG | LFG |
| Basic Sentences | 1 (8) | 2 (7) | 3(9) |
| Auxiliary Verbs | 3 (9) | 4 (8) | 2 (9) |
| Case Assigning PPs | 4 (7) | 3 (8) | 4 (6) |
| Adpositional Sentences | 4 (6) | 3 (9) | 4 (7) |
| Descriptive Adjectives | 3 (8) | 2 (7) | 4 (7) |
| Genitive Case | 2 (9) | 3 (7) | 4 (6) |
| Predictive Adjectives | 2 (8) | 3 (7) | 4 (7) |
| Relative Clause | 5 (8) | 4 (7) | 5 (8) |

Table 2: Highest Voted Ranks by Grammar Writers for each grammar

## 4. RESULTS

We calculated the results for total and average time taken to complete the task, accuracy with which the task was completed, difficulty ratings provided by each writer, for each formalism, on different categories of sentences. The types of errors committed. The results of the study are provided in the following sections.

### 4.1 Time Taken and Accuracy

Table 1 summarizes the average time taken to complete the task by each writer and the accuracy with which they did it. Looking at the data, it is clearly seen that time taken to complete the task was least in PG and most in LFG, TAG was in between the two.

The average time taken by the writers to complete the task for PG, TAG and LFG is 10.3 min, 34 min and 56 min respectively. PG took least time with which the sentences were completed. We also measured average accuracy of each writer. Here TAG scored more accuracy then the other formalisms.

### 4.2 Difficulty Rating

After the task, we provided the questionnaire to the writers. We asked them to provide us with the difficulty rating for each type of sentence, for each grammar. We asked them to ranks the difficulty of sentences between 1 and 5, where 1 being the easiest and 5 being the toughest. As it was not possible to provide results for all the writes here. In Table 2, we provide the ranks given majority of writers. The sores without brackets are the ranks given and the ones in bracket are the no. of writers who gave this rank.

We can see that PG sores very well in simple, generative and predictive adjective cases, but do not perform well on other categories of the sentences. TAG on the other hand performs moderately well. It sores highest in four categories of sentences. LFG scores highest in just one.

### 4.3 Error Analysis

We also examined the types of errors committed by different formalisms, as we wanted to know, why writer had great difficulty with LFG as compared to PG or TAG.

In Paninian Grammar, we analyzed that writers faced great difficulty in assigning relationships to dependency structures. This could be due to the notational convention of the formalism or due to the difficulty with the concepts of head and dependents. We also saw that whenever a directional error was made, the correct rule was framed for implementing the dependency. This shows that the difficulty was with the notion and not with the concepts. Moreover PG although being least rigorous out of the three, showed some sluggishness while dealing with complex sentences.

In Tree Adjoining Grammar, we saw that most errors were in the formation of the derived trees. Between adjunction and substitution operations, adjunction proved to be more error prone. Almost 80% errors were made due to incorrect adjunction operation. This shows that writers had great difficulty understanding the adjunction operation.

In Lexical Functional Grammar, we saw that grammar writer's had great difficulty in associating features with constituent structures. In some cases the writers got confused as to use noun phrase and verb phrase in constituent structure or to use subject and predicate in feature structure. Although this formalism is the most perfect in terms of linguistic phenomena as it captures all the aspects of the language's grammar, it is also fairly difficulty to understand, as it takes time for the grammar write to understand and implement grammar using it.

## 5. CONCLUSION

We wanted to study the applicability of different grammar formalism on Indian languages, so that different NLP tasks like development of a deep probabilistic parser or development of a Treebank could be under taken. In doing so, we gathered insights into the different formalisms and understood the merits and demerits of each.

We found out that though Paninain Grammar was preferred for simple sentences, overall performance of TAG was good. It scored better in the average accuracy attained to write the sentences. Although this is a preliminary study and more in-depth evaluations are required before making any sound conclusions. But, with some confidence we can say that TAG can perform better in most of the difficult cases as compared to dependency grammar.

## 6. REFERENCES

[1]. J.G. Neal, E.L. Feit and C.A. Montgomery, "Benchmark Investigation/Identification Project," Machine Translation, Springer, Germany, Vol 8, No. 1-2, pp77-84, 1993.
[2]. T. Baldwin, J. Beavers, E.M. Bender, D. Flickinger, A. Kim, and S. Oepen, "Beauty and the beast: What running a broad-coverage precision grammar over thee bnc taught us about the grammar — and the corpus," Linguistic Evidence: Empirical, Theoretical, and Computational Perspectives, Mouton de Gruyter, Berlin, Germany, pp 49–70., 2005.
[3]. S. A. Waterman, "Distributed parse mining," In Proceedings of the NAACL Workshop on Software Engineering, Testing, and Quality Assurance for Natural Language Processing, USA, 2009.
[4]. C.D. Manning and H Schütze, "Foundations of Statistical Natural Language Processing", MIT Press, USA, 1999.
[5]. E. Charniak, "A Maximum Entropy Inspired Parser", In Proceedings of NAACL, USA, 2000.
[6]. J. Nivre, J. Hall and J. Nilsson, "MaltParser: A Data-Driven Parser-Generator for Dependency Parsing," In Proceedings of the fifth international conference on Language Resources and Evaluation, Genoa, Italy, pp. 2216-2219, May, 2006.
[7]. A Taylor, A. Warner and B. Santorini, "The Penn Treebank: An Overview", Treebanks: Building and Using Parsed Corpora, Kluwer Academic Publishers, Netherlands, 2003.
[8]. S. Brants, S. Dipper, P. Eisenberg, S. Hansen, E. König, W. Lezius, C. Rohrer, G. Smith, and H. Uszkoreit, "TIGER: Linguistic interpretation of a German corpus," Research on Language and Computation, Springer, Germany, Vol 9, No. 2, pp 597-620, 2004.
[9]. J Hajič, B Hladká and P. Pajas, " The Prague Dependency Treebank: Annotation Structure and Support," In Proceedings of the IRCS Workshop on Linguistic Databases, Pennsylvania, USA, pp 105-114, 2001.
[10]. M. Covington, "Parsing Discontinuous Constituents Dependency in Dependency Grammar," Computational Linguistics, MIT Press, USA, Vol 16, No. 4, pp-234-236, 1990
[11]. R. Sangal and V Chaitanya, "An Intermediate Language for Machine Translation: An approach based on Sanskrit using conceptual graph notation", Journal of Computer Society of India, Mumbai, India, Vol 17, pp 9-21, 1987.
[12]. A.K. Joshi, "An Introduction to Tree Adjoining Grammars," Mathematics of Language, John Benjamins, Netherlands, 1987.
[13]. I.A. Sag, T Wasow and E.M. Bender, "Syntactic Theory," 2 Edition, CSLI Publications, USA, 2001.
[14]. M. Dalrymple, "Lexical Functional Grammar: Syntax and Semantics", Academic Press, USA, 2001.
[15]. H. Gaifman, "Dependency systems and phrase structure systems," Information and Control, USA, Vol 8, pp 304-337, 1965.
[16]. J. Hudson, "Word Grammar," Basil Blackwell, England, 1984.
[17]. S. Starosta, "The Case for Lexicase: An Outline of Lexicase Grammatical Theory," Cassell, London, 1988.
[18]. A. Bharti, V. Chaitanya, R. Sangal, "Natural Language Processing: A Paninian Perspective," PHI, India, 1999.
[19]. V. Dwivedi, "Tropicalization in Hindi and the Correlative Construction," Theoretical Perspectives on Word Order in South Asian Languages, CSLI Publications, USA, 1994.